\documentclass[letterpaper, 10 pt, conference]{ieeeconf}  

\IEEEoverridecommandlockouts                              

\usepackage{cite}
\usepackage{amsmath,amssymb,amsfonts}
\usepackage{algorithmic}
\usepackage{booktabs}
\usepackage{graphicx}
\usepackage{textcomp}
\usepackage{xcolor}
\usepackage{etoolbox}
\usepackage{graphicx}
\usepackage[skip=0.333\baselineskip]{caption}
\usepackage{subcaption}
\usepackage{multirow}
\usepackage{multicol}
\usepackage{tabularx}
\usepackage{hyperref}

\title{\LARGE \bf
FADet: A Multi-sensor 3D Object Detection Network based on Local Featured Attention
}

\author{
    Ziang Guo \and 
    Zakhar Yagudin \and 
    Selamawit Asfaw \and 
    Artem Lykov \and 
    Dzmitry Tsetserukou
    \thanks{The authors are with the Intelligent Space Robotics Laboratory, Center for Digital Engineering, Skolkovo Institute of Science and Technology, Moscow, Russia
    \tt \{ziang.guo, Zakhar.Yagudin, Selamawit.Asfaw, artem.lykov, d.tsetserukou\}@skoltech.ru}
}

\begin{document}

\maketitle
\thispagestyle{empty}
\pagestyle{empty}

\begin{abstract}
Camera, LiDAR and radar are common perception sensors for autonomous driving tasks. Robust prediction of 3D object detection is optimally based on the fusion of these sensors. To exploit their abilities wisely remains a challenge because each of these sensors has its own characteristics. In this paper, we propose FADet, a multi-sensor 3D detection network, which specifically studies the characteristics of different sensors based on our local featured attention modules. For camera images, we propose dual-attention-based sub-module. For LiDAR point clouds, triple-attention-based sub-module is utilized while mixed-attention-based sub-module is applied for features of radar points. With local featured attention sub-modules, our FADet has effective detection results in long-tail and complex scenes from camera, LiDAR and radar input. On NuScenes validation dataset, FADet achieves state-of-the-art performance on LiDAR-camera object detection tasks with $71.8\%$ NDS and $69.0\%$ mAP, at the same time, on radar-camera object detection tasks with $51.7\%$ NDS and $40.3\%$ mAP. Code will be released at \href{https://github.com/ZionGo6/FADet}{\textit{https://github.com/ZionGo6/FADet}}.
\end{abstract}

\section{Introduction}

\subsection{Motivation}

Perception based on input from multi-sensor, such as cameras, radar, and LiDAR (light detection and ranging) is essential in the development
 of autonomous driving system. Cameras can capture richer features than LiDARs and radars with texture information while LiDARs and Radars are able to provide more geometric information. Utilizing the features from different sensors effectively empowers the perception system in autonomous driving and robotics to achieve robust and accurate performance \cite{yan2023CMT}. \par However, data input from different sensors normally is with its corresponding representations, which leads to less homogeneity of extracted features. We consider working on dealing with queries from different modalities. For instance, when dealing with camera image input, query is the extracted input that the model is trying to match with the most relevant features from the image, if a model is trying to detect certain objects in the image, the query can be the features representing the object itself \cite{wu2020query}. Besides, finding correlated features based on multiple modalities through global attention mechanism leads to heavy computational cost, thus resulting in misalignment of data processing from dynamic driving scenes \cite{Zhang2023AFTR}. When having multi-modal data as input, it is necessary to keep the attention module focused on corresponding extracted features based on the attraction from different modalities, potentially benefiting the detection accuracy and inference latency. \par
\begin{figure}[t]
\includegraphics[width=0.45\textwidth,height=0.5\textheight, keepaspectratio]{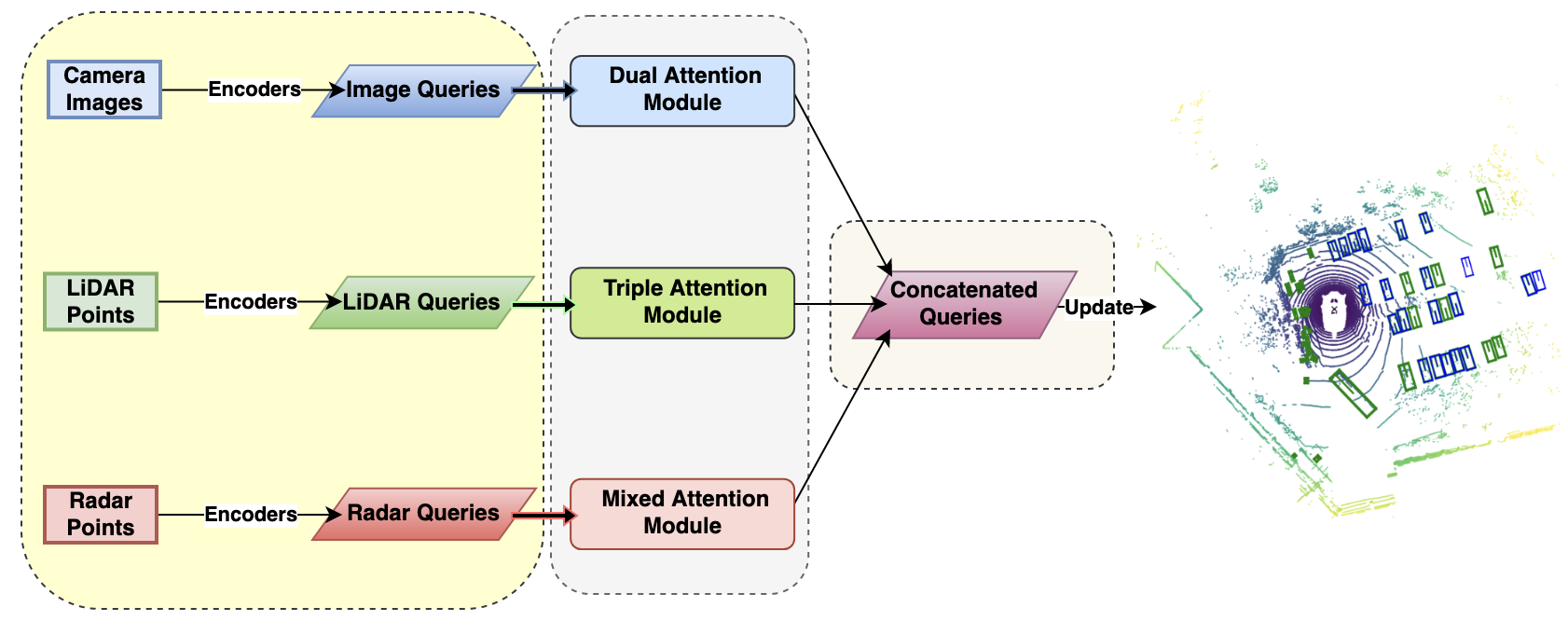}
\caption{Illustration of our FADet layout. Sensor data is encoded and represented as queries. Then our local featured attention sub-modules process queries to provide feature-specific weights for feature concatenation.}
\label{fig:whole}
\end{figure}

 To address these problems, we propose FADet, a camera-LiDAR-radar 3D object detection network working through local featured attention mechanism which is customized based on the characteristics of different modalities. Similar to FUTR3D \cite{chen2023futr3d}, features are encoded from different modalities and then are performed fusion based on feature sampling. Query features from different modalities are input to our customized local attention modules. The obtained attention weights are aggregated with feature maps of different data towards the output of multi-head attention module. Through paying specific attention to different modalities on a local scale, we want to improve the performance of global attention and thus boost the corresponding detection results. The whole scheme is demonstrated in Fig. \ref{fig:whole}.
 
\subsection{Related Work}

\subsubsection{Camera-based Detection Methods}

Y. Liu et al. \cite{liu2022star} introduce Star-Convolution module to enhance detection accuracy in both monocular and stereo scenarios. By embedding camera projection knowledge in neural networks which allows for end-to-end learning, the performance may be affected by initial bounding box proposals. SparseBEV \cite{liu2023sparsebev} introduces a fully sparse 3D object detector that focuses on enhancing 3D object detection performance in Bird's-eye view (BEV) space, particularly in multi-camera videos. By proposing adaptive self-attention, spatio-temporal sampling, and adaptive mixing techniques, SparseBEV achieves state-of-the-art performance on the nuScenes dataset. However, sparse detectors in comparison to dense detectors resulted in less reliability in complex scenarios. Huang et al. \cite{huang2022bevdet4d} introduce BEVDet4D, a novel paradigm for enhancing 3D object detection by incorporating temporal cues and feature fusion techniques. However, The limitations of BEVDet4D are posed by the complexities of directly concatenating features.

\subsubsection{LiDAR-based Detection Methods}

Hu et al. \cite{hu2022pointdense} introduce the Point Density-Aware Voxel network (PDV) for LiDAR 3D object detection. By leveraging kernel density estimation and self-attention for feature aggregation and refining bounding box confidences based on LiDAR point density, the PDV method significantly enhances 3D object detection for autonomous driving systems. However, the study lacks a detailed analysis of the computational complexity of the proposed method, potentially impacting practical implementation and real-time performance. You et al. \cite{you2019pseudo} present significant contributions to improve the depth estimation of faraway objects within the pseudo-LiDAR framework using an enhanced stereo network architecture, a refined loss function and a depth propagation algorithm for diffusing sparse LiDAR measurements. However, despite the improvements, the reliance on sparse LiDAR data may still pose challenges in capturing comprehensive 3D detection information. Shi et al. \cite{shi2023center} present a work of utilizing BEV feature map for self-attention and cross-attention blocks to employ point-voxel feature set abstraction for 3D object detection, which enhances the detection process by incorporating dense supervision for center candidate generation from nearby pixels.

\subsubsection{Multi-modal Detection Methods}

Kim et al. \cite{kim2023rcm} demonstrate state-of-the-art performance through the fusion of radar and camera features at both feature and instance levels leveraging radar guided BEV encoders for feature-level fusion and radar grid point refinement module for instance-level fusion. Nonetheless, inherent characteristics of radar point clouds may impact the precision in certain scenarios. Zhao et al. \cite{zhao2024bevradar} a innovative bidirectional radar-camera fusion technique. Through the integration of a bidirectional spatial fusion module and a promoted detection transformer, the model excels in unifying feature representations from distinct domains. Wang et al. \cite{wang2022m2} revolutionize point cloud feature fusion by seamlessly integrating radar and LiDAR data sources. However, a limitation of this study may pertain to the scalability and adaptability across different vehicle types. Chen et al. \cite{chen2023futr3d} introduce FUTR3D, a groundbreaking unified sensor fusion framework that integrates diverse sensor configurations, including LiDARs, cameras and radars. Despite great adaptability across various sensor types, FUTR3D lacks in-depth study on features of different sensor data.

\subsection{Contribution}

Our main contributions are as follows. For LiDAR point cloud input, considering the three-dimensional features of LiDAR points, we propose a sub-module named triple attention for processing LiDAR points. For camera image input, a sub-module named dual attention is implemented based on characteristics of image. For radar, we integrate convolution and self-attention as a sub-module to process radar points. Analyzing both the spatial and channel-wise features from different modalities on a local scale, our FADet is able to detect critical long-tail objects and complex objects achieving state-of-the-art 3D detection performance on nuScenes dataset on both LiDAR-camera and radar-camera detection tasks \cite{nuscenes}.

\section{Framework Overview}

\subsection{Encoders}

For surrounding camera image input based on nuScenes dataset \cite{nuscenes}, our framework utilizes VoVNet \cite{park2021VoVNet} and Recursive Feature Pyramid (RFP) \cite{qiao2021RFP} for image feature extraction, where the feature maps are output through RFP neck. Thus, we have 
$\mathbf{F}_{\mathrm{Camera}}^{ki} \in \mathbb{R}^{C \times H_i \times W_i}$ for the $k$-th camera, 
where $C$ is the number of output channels and $H_i \times W_i$ is the size of the $i$-th feature map as feature encoder output. About LiDAR point cloud, VoxelNet \cite{zhou2018voxelnet} and Feature Pyramid Network (FPN) \cite{lin2017FPN} are used as Bird’s-eye view (BEV) feature encoders. The obtained feature map of point cloud is denoted as 
$\mathbf{F}_{\mathrm{LiDAR}}^i \in \mathbb{R}^{C \times H_i \times W_i}$, 
where $C$ is the number of output channels and $H_i \times W_i$ is the size of the $i$-th feature map. In terms of radar points, the pillar features of radar points are first extracted and BEV feature maps are obtained by pillar feature network \cite{lang2019pointpillars}. Then we obtain radar feature map as 
$\mathbf{F}_{\mathrm{radar}} \in \mathbb{R}^{C_{n} \times H \times W}$, 
where $C_n$ is the batch of radar points and $H \times W$ is the size of the feature map of each batch. Via our encodes, regions of interest from camera images, LiDAR point clouds and radar points are identified and prepared for further processing \cite{hossain2023ROI}.

\subsection{Attention}

Following deformable DETR \cite{zhu2020deformableDETR}, the queries are first randomized with reference points that are relative coordinates in three dimensions. As $\mathbf{F}_{\mathrm{Camera}}^{ki}$ is the image feature from $k$-th camera after encoders, the reference points are then projected by the intrinsic and extrinsic matrix of the camera and are denoted as $\mathcal{P}^c_k(ref)$. Attention weight $\mathbf{A}_{\mathrm{Camera}}^{ki}$ is obtained from query $ Q_i \in \mathbb{R}^{N_q \times H'_i \times W'_i}$, 
where $N_q$ is the number of queries and $H'_i \times W'_i$ is the size of the $i$-th feature map. Proposed dual attention module relates semantic information in spatial and channel dimensions respectively, which potentially contributes to more precise image pixel classification \cite{fu2019dual}. Via dual attention module, obtained attention weights $\mathbf{A}_{\mathrm{Camera}}^{ki}$ carry both channel and spatial attention. Then the output image feature is computed as follows:
\begin{equation*}
   \mathbf{F'}_{\mathrm{Camera}}^i = \sum_{k=1}^N \sum_{i=1}^m \mathbf{F}_{\mathrm{Camera}}^{ki} (\mathcal{P}^c_k(ref)) \cdot \mathbf{A}_{\mathrm{Camera}}^{ki}\ ,    \tag{1}
\end{equation*}
\newline where $N$ is the number of surrounding cameras and $m$ means the number of feature maps.

\par LiDAR point cloud feature after encoders $\mathbf{F}_{\mathrm{LiDAR}}^i \in \mathbb{R}^{C \times H_i \times W_i}$ is thus processed by a triple attention module as follows:
\begin{equation*}
    \mathbf{F'}_{\mathrm{LiDAR}}^i = \sum_{i=1}^m \sum_{k=1}^K \mathbf{F}_{\mathrm{LiDAR}}^i (\mathcal{P}^L_k(ref) + \Delta_{\mathrm{LiDAR}}^{ki}) \cdot \mathbf{A}_{\mathrm{LiDAR}}^{ki} \ , \tag{2}
\end{equation*}
\newline where $\mathbf{F'}_{\mathrm{LiDAR}}^i$ is the output feature from $i$-th feature map, $m$ means the number of feature maps, $K$ means the sampled points from each feature map, $\mathcal{P}^L_k(ref)$ is the projection from reference point in three dimensions to BEV map. $\Delta_{\mathrm{LiDAR}}^{ki}$ means the sampling offset between reference points and points in feature maps and $\mathbf{A}_{\mathrm{LiDAR}}^{ki}$ is the attention weight induced by triple attention module. \par Implemented triple attention adapts input query feature by bridging the connections among input channels \cite{misra2021triple}. For instance, the input query is as $ Q_i \in \mathbb{R}^{N_q \times H'_i \times W'_i}$, where $N_q$ is the number of queries and $H'_i \times W'_i$ is the size of the $i$-th feature map. Triple attention works on three branches, which are the computation between channel dimension $N_q$ and spatial dimension $H'_i$, channel dimension $N_q$ and spatial dimension $W'_i$, spatial dimensions $H'_i$ and $W'_i$. Then attention weights are operated by adaptive max pooling, convolution layer, batch normalization and sigmoid activation layer and then are aggregated from all three branches.
\par To deal with radar points, we have radar features after encoders $\mathbf{F}_{\mathrm{radar}} \in \mathbb{R}^{C_{n} \times H \times W} $, 
where $C_n$ is the batch of radar points and $H \times W$ is the size of the feature map of each batch. Then the sampled radar feature with mixed attention weights is obtained as follows:
\begin{equation*}
        \mathbf{F'}_{\mathrm{radar} }^i = \sum_{k=1}^K \mathbf{F}_{\mathrm{radar}}( \mathcal{P}^r_k(ref) + \Delta_{\mathrm{radar}}^{ki} ) \cdot \mathbf{A}_{\mathrm{radar}}^{ki}\ ,     \tag{3}
\end{equation*}
\newline where $\mathcal{P}^r_k(ref)$ is the projection from the reference point to BEV map and $\Delta_{\mathrm{radar}}^{ki}$ is the sampling offset between reference points and points in feature maps, and $\mathbf{A}_{\mathrm{radar}}^{ki}$ is the attention weights processed by mixed attention module. Input query features are passed towards both convolution-based and self-attention-based branches, where convolution-based branch extracts feature with a $3 \times 3$ kernel and self-attention-based branch processes feature via queries, keys and values from three $1 \times 1$ convolution kernels. \par Finally, the output features from each local featured attention modules are concatenated and encoded by multi-layered perceptron network and queries are updated with position encoding by feed forward network.

\subsection{Decoder}

Decoder in our framework processes queries in parallel at each decoder layer and predicts the three-dimensional bounding boxes with box center coordinate $\mathcal{X}_i \in  \mathbb{R}^{x, y, z}$, box size ($w_i, h_i, l_i$), orientation ($\sin{\theta}_i, \cos{\theta}_i$), velocity $v_i \in \mathbb{R}^{V_x, V_y}$ \cite{carion2020detr}.

\subsection{Loss Functions}

Following DETR3D \cite{wang2022detr3d}, loss functions are the weighted sum of adapted classification loss, regression L1 loss and regression IoU loss as 0.7, 0.2 and 0.1, respectively \cite{Lin2017FocalLoss}.

\section{Methodology}

\subsection{Dual Attention for Image Pixel}

Considering image pixel characteristics, they can be categorized into two main types: channel-wise features and spatial-wise features, where channel-wise features denote statistical color or gray-scale description of pixels, while spatial-wise features capture the spatial arrangement and visual patterns of objects within the image \cite{liu2022pscc}. \par Proposed dual attention module provides both channel-aware and spatial-aware concern. The input encoded query $Q^{Camera}_i \in \mathbb{R}^{N_{q} \times H_i \times W_i}$, where $N_q$ is the number of queries and $H_i \times W_i$ is the size of the $i$-th feature map from images is passed through convolution layer with $3 \times 3$ kernel to generate two middle feature maps $Q_{1i}$ and $Q_{2i} \in \mathbb{R}^{N_{q} \times H_i \times W_i}$. Then matrix multiplication is performed between $Q_{1i}$ and the transpose of $Q_{2i}$. Thus, the spatial attention map $\mathbf{S}_{mn} \in \mathbb{R}^{(H_i \times W_i) \times (H_i \times W_i)}$ is computed as follows:

\begin{equation}
\mathbf{S}_{mn}=\frac{exp(Q_{1m}\cdot Q_{2n})}{\sum_{m,n=1}^{H_i \times W_i}exp(Q_{1m}\cdot Q_{2n})}\ ,   \tag{4}
\end{equation}
where $m,n$ are the elements in the feature map $H_i \times W_i$.
\par The input encoded query $Q^{Camera}_i \in \mathbb{R}^{N_{q} \times H_i \times W_i}$ is also passed through another convolution layer with $3 \times 3$ kernel to generate middle feature map $Q_{3i} \in \mathbb{R}^{N_{q} \times H_i \times W_i}$. Subsequently, matrix multiplication is performed between the transpose of $\mathbf{S}_{mn}$ and $Q_{3i}$, producing the final output of spatial-aware operations:

\begin{equation}
\mathbf{S}_{n} = \alpha \sum_{m,n=1}^{H_i \times W_i}(trans(\mathbf{S}_{mn})Q_{3m} ) + Q^{Camera}_{n}\ ,     \tag{5}
\end{equation}
where $\alpha$ is the learnable weight.
\par In channel-aware branch, channel attention map $\mathbf{C}_{mn} \in \mathbb{R}^{N_q \times N_q}$ is given by:

\begin{equation}
\mathbf{C}_{mn} = \frac{exp(Q^{Camera}_m\cdot Q^{Camera}_n)}{\sum_{m,n=1}^{N_q}exp(Q^{Camera}_m\cdot Q^{Camera}_n)}\ , \tag{6}
\end{equation}
\par Afterwards, matrix multiplication is performed between the transpose of $\mathbf{C}_{mn}$ and $Q^{Camera}_i$. Then the final output of channel-aware operations is as follows:

\begin{equation}
\mathbf{C}_{n} =  \beta \sum_{m,n=1}^{N_q}(trans(\mathbf{C}_{mn})Q^{Camera}_m ) + Q^{Camera}_n\ ,    \tag{7}
\end{equation}
where $\beta$ is the learnable weight.
\par
\begin{figure}
\centering
    \begin{subfigure}{.47\linewidth}
        \centering
        \includegraphics[height=0.38\textheight, keepaspectratio]{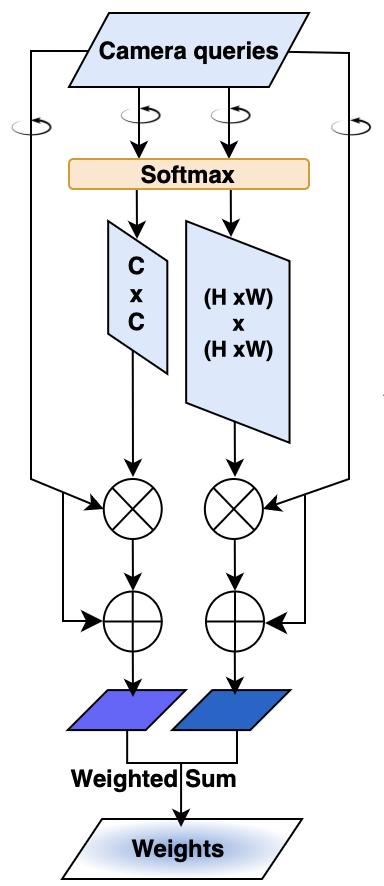}
        \caption{}
    \end{subfigure}
    \begin{subfigure}{.47\linewidth}
    \centering
    \includegraphics[height=0.39\textheight, keepaspectratio]{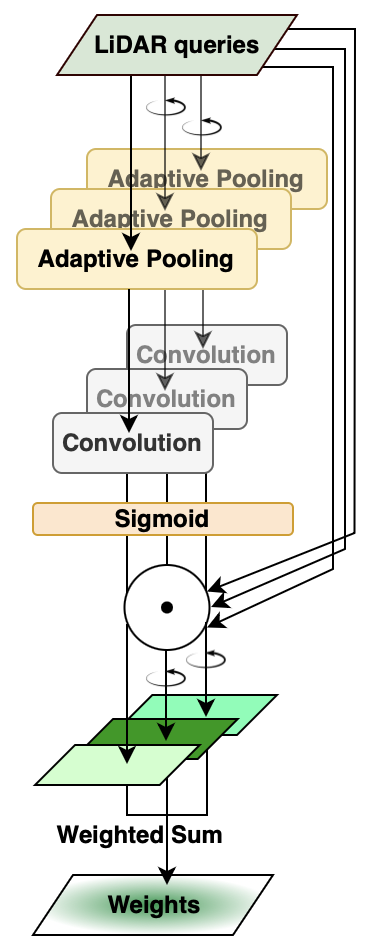}
    \caption{}
    \end{subfigure}

\caption{Illustrations of camera and LiDAR local attention modulues. (a) Layout of local dual attention module for camera features, where $\bigoplus$ means element-wise summation and $\bigotimes$ means matrix multiplication. (b) Layout of local triple attention module for LiDAR features, where $\bigodot$ means element-wise product.}
\label{fig:camera+lidar}
\end{figure}

 The output attention weights of dual attention module is thus $\mathbf{A}_{\mathrm{Camera}}^{ki} = \mathbf{S}_{i} + \mathbf{C}_{i} $. The scheme is shown in Fig. \ref{fig:camera+lidar}.

\subsection{Triple Attention for LiDAR Point Cloud}

LiDAR point cloud is a collection of three-dimensional points that provides detailed information about the surface of an object. Each point in the cloud represents the exact location of a laser pulse that has bounced off a surface and been measured by the LiDAR sensor \cite{elharrouss2023LiDAR}. Considering such features of LiDAR point cloud, we propose triple attention to process the attention weights obtained from LiDAR queries. \par The scheme of triple attention modules is in Fig. \ref{fig:camera+lidar}, where firstly the input encoded query $Q^{LiDAR}_i \in \mathbb{R}^{N_{q} \times H_i \times W_i}$, 
where $N_q$ is the number of queries and $H_i \times W_i$ is the size of the $i$-th feature map from LiDAR point cloud features is conveyed to adaptive max pooling with the size of $H_i \times W_i$ and then a convolution layer with $5 \times 5$ kernel. Similarly, in the second branch, the input encoded query is permuted as $Q'^{LiDAR}_i \in \mathbb{R}^{H_i \times N_q \times W_i} $, while the same operation as first branch is performed between $N_q \times W_i$ with respect to $H_i$. In the third branch, operations are performed between $N_q \times H_i$ with respect to $W_i$. Activated by a sigmoid function, permuting towards the original input shape, attention weights from three branches $\mathbf{A}_{\mathrm{LiDAR}}^{ki}$ are output as a weighted sum. The procedure is defined by:

\begin{equation*}
    \mathcal{X} = AdaptivePooling_{H_i \times W_i}(Q^{LiDAR}_i \in \mathbb{R}^{N_{q} \times H_i \times W_i})      
\end{equation*}
\begin{equation*}
    \Tilde{\mathcal{X}} = Convolution_{5 \times 5}(\mathcal{X})      
\end{equation*}
\begin{equation*}
    \Tilde{\mathcal{A}} = Permute(Sigmoid(\Tilde{\mathcal{X}}))      
\end{equation*}
\begin{equation*}
    \mathbf{A}_{\mathrm{LiDAR}}^{ki} = \alpha_1 \Tilde{\mathcal{A}} + \alpha_2 \Tilde{\mathcal{A'}} + \alpha_3 \Tilde{\mathcal{A''}}\ ,      \tag{8}
\end{equation*}
where $\mathbf{A}_{\mathrm{LiDAR}}^{ki}$ is weighted summed from three branches, $\Tilde{\mathcal{A'}}$ and $\Tilde{\mathcal{A''}}$ are the weights obtained from permuted $Q^{LiDAR}_i$. 

\subsection{Mixed Attention for Radar Point}

\begin{figure}
\centering
\includegraphics[width=0.4\textwidth,height=\textheight, keepaspectratio]{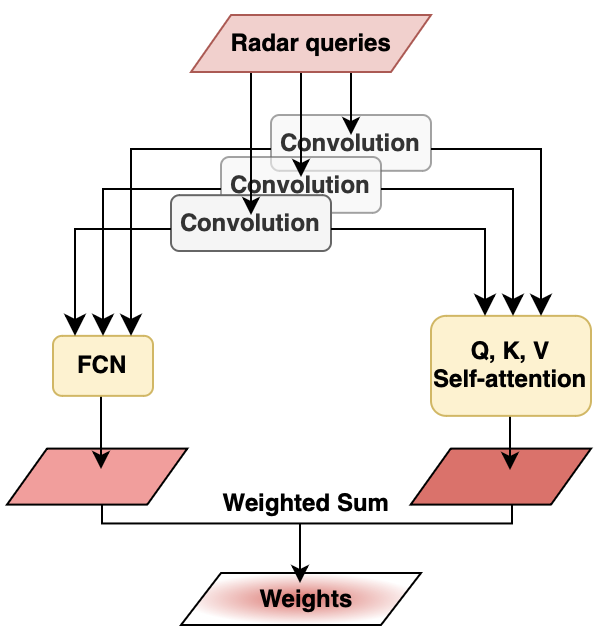}
\caption{Layout of local mixed attention module for radar features.}
\label{fig:radar}
\end{figure}

For radar points, we use a combination of convolution and self-attention operations to process radar data. When we have input encoded query $Q^{radar}_i \in \mathbb{R}^{N_q \times H_i \times W_i}$, 
where $N_q$ is the number of queries and $H_i \times W_i$ is the size of the $i$-th feature map, three convolution kernels with the size of $1 \times 1$ are used to generate $3 \times N_{Mix}$ middle feature maps, where $N_{Mix}$ is the number of heads of mixed attention module. Then the convolution branch will perform full connection with the kernel size of $1 \times 1$, while the attention branch takes $3 \times N_{Mix}$ middle feature maps as query-key-value use to calculate the attention weights. Finally, the weights from two branches are performed weighted sum as $\mathbf{A}_{\mathrm{radar}}^{ki}$. The procedure is denoted as follows:

\begin{equation*}
    \mathcal{X}_1,  \mathcal{X}_2, \mathcal{X}_3 = 3 \times Convolution_{1 \times 1}(Q^{radar}_i \in \mathbb{R}^{N_q \times H_i \times W_i})
\end{equation*}
\begin{equation*}
    \mathcal{A}_{conv} = FCN(\mathcal{X}_1,  \mathcal{X}_2, \mathcal{X}_3)
\end{equation*}
\begin{equation*}
    \mathcal{A}_{attention} = Attention(\mathcal{X}_1,  \mathcal{X}_2, \mathcal{X}_3)
\end{equation*}
\begin{equation*}
    \mathbf{A}_{\mathrm{radar}}^{ki} = \alpha_{conv}\mathcal{A}_{conv} + \alpha_{attention}\mathcal{A}_{attention} \ ,                    \tag{9}
\end{equation*}
where $FCN$ is the operation of fully connected layer, $Attention$ is the self-attention-like operation. $\mathbf{A}_{\mathrm{radar}}^{ki}$ is calculated as weighted sum. Fig. \ref{fig:radar} presents the layout of the mixed attention module for radar features.

\section{Experiments}

\subsection{Dataset}

nuScenes dataset \cite{nuscenes} is chosen for training and evaluation of our framework, which is a large-scale dataset for autonomous driving, consisting of 850 scenes with measurements from 1 spinning LiDAR, 6 cameras and 5 long-range radars, etc. LiDAR is with 20 Hz capture frequency and 32 beams, while cameras are with 12 Hz capture frequency and $1600 \times  900$ resolution. Radars are with 13 Hz capture frequency. Well synchronized keyframes from cameras, LiDAR and radars are at 2 Hz.

\subsection{Experimental Setup}

The feature dimension $C$ is set as $256$ for LiDAR feature $\mathbf{F}_{\mathrm{LiDAR}}^i \in \mathbb{R}^{C \times H_i \times W_i}$, image feature $\mathbf{F}_{\mathrm{Camera}}^{ki} \in \mathbb{R}^{C \times H_i \times W_i}$. For radar points $\mathbf{F}_{\mathrm{radar}} \in \mathbb{R}^{C_{n} \times H \times W}$, $C_n = 64$. $N_q = 900$ for object queries. For dual attention, we use kernel size $3 \times 3$ for generating middle feature maps and $5 \times 5$ kernels are used in triple attention after adaptive pooling. In the mixed attention module, $1 \times 1$ kernels are used for convolution operations. We trained both LiDAR-camera and radar-camera models 24 epochs using AdamW optimizer \cite{loshchilov2017AdamW} with a learning rate at $1.0 \times 10^{-4}$ on a single Nvidia RTX 4090 24G on nuScenes dataset \cite{nuscenes}.

\subsection{Evaluation Methods}

We show Mean Average Precision (mAP) and nuScenes Detection Score (NDS) as our experimental metrics. Mean Average Precision (mAP) means the distances between the centers of ground truth and predicted bounding boxes in bird’s-eye view. NDS means the errors of detection results by considering these errors: Average Translation Error (ATE), Average Scale Error (ASE), Average Orientation Error (AOE), Average Velocity Error (AVE), and Average Attribute Error (AAE).

\subsection{Comparison with Other State-of-the-Art Methods}

\begin{figure*}[h]
\centering
        \includegraphics[width=0.9\linewidth]{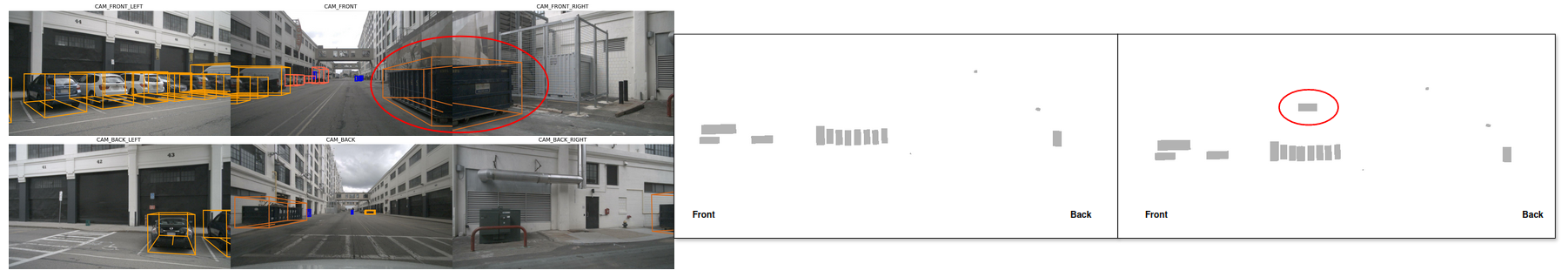} \\
        \includegraphics[width=0.9\linewidth]{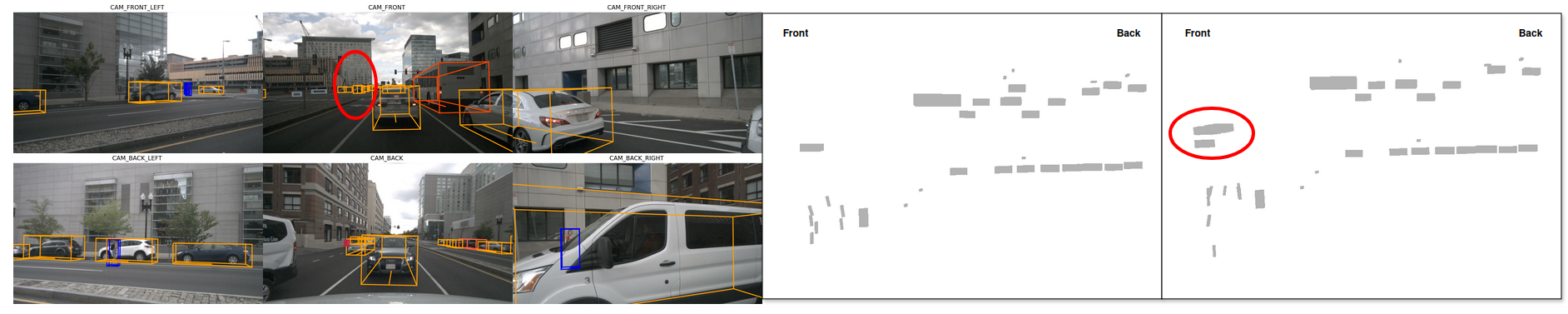} \\
        \includegraphics[width=0.9\linewidth]{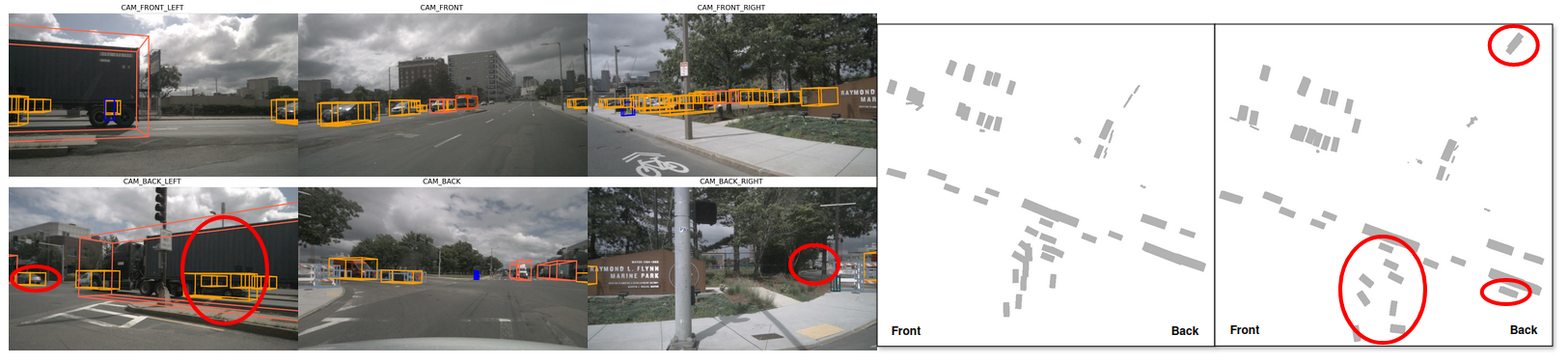} \\
        \includegraphics[width=0.9\linewidth]{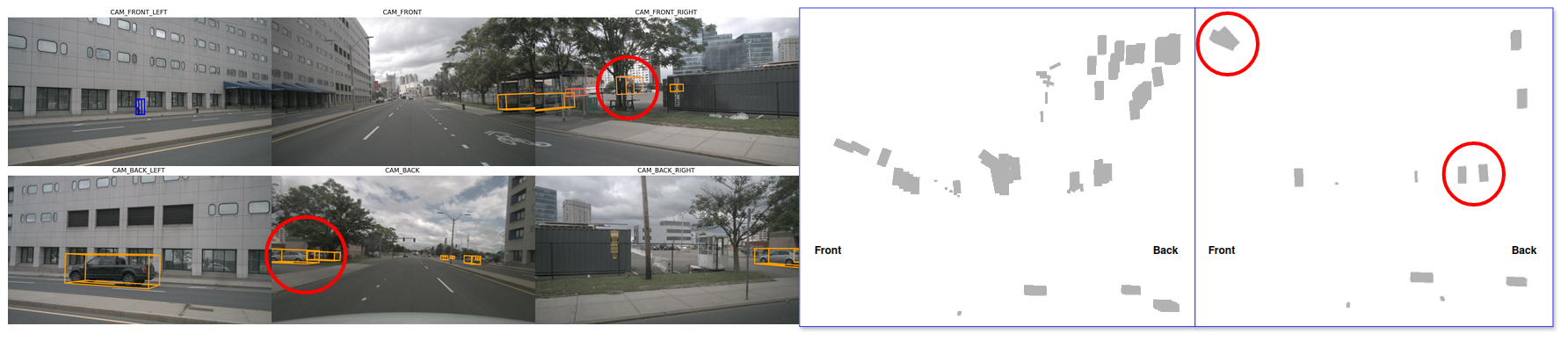} \\
        \includegraphics[width=0.9\linewidth]{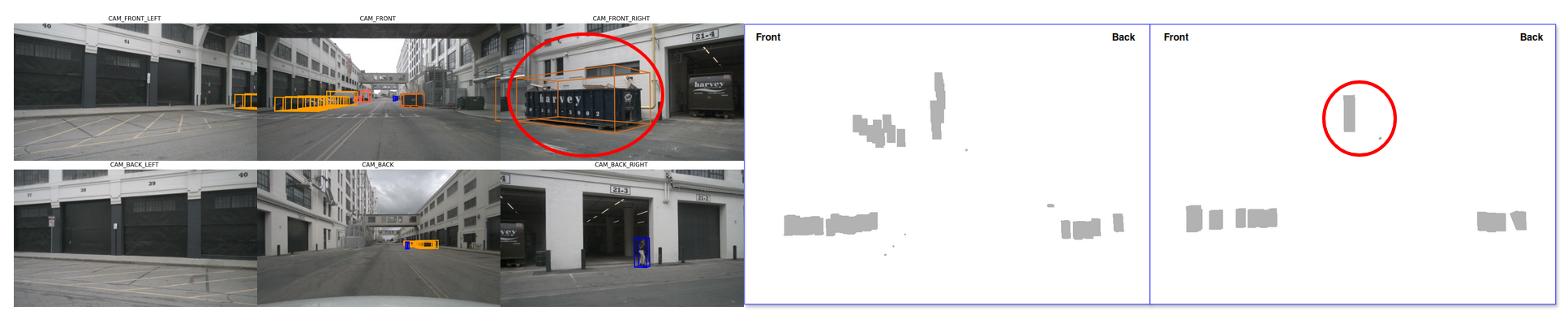} \\
        \caption{Above three sets of figures: (Left) The surrounding views of 6 cameras with ground truth annotations from nuScenes dataset. (Middle) The prediction results of FUTR3D on \textbf{LiDAR-camera} tasks. (Right) The prediction results of our FADet on \textbf{LiDAR-camera} tasks. Below two sets of figures are corresponding results on \textbf{radar-camera} tasks. From the marked red circles on our FADet prediction and surrounding images, we can see the corresponding detection in long-tail and complex cases.}
\label{fig:teasers}
\end{figure*}

In Table \ref{table:comparison}, we provide quantitative results with nuScenes Detection Score (NDS) and Mean Average Precision (mAP) comparison with other state-of-the-art methods on nuScenes validation set. Our FADet achieves $71.8\%$ NDS and $69.0\%$ mAP on LiDAR-camera detection tasks, while $51.7\%$ NDS and $40.3\%$ mAP on radar-camera detection tasks. Besides, based on visualization comparison with FUTR3D on more than 6000 scenes of nuScenes dataset, our FADet also shows improvement in long-tail and complex cases with critical detection. Part of the visualization results on LiDAR-camera experiments are shown in Fig. \ref{fig:teasers}.

\begin{table}
    \caption {NDS/mAP (\%) comparison on nuScenes validation set.}
    \label{table:comparison}
    \centering
    \begin{tabular}{c|cc}
        \hline\noalign{\smallskip}
        \multirow{2}{*}{Models} & \multicolumn{2}{c}{Modalities} \\
         & LiDAR + Camera & Radar + Camera \\
        \midrule
        TransFusion \cite{bai2022transfusion} & 71.3 / 67.5 & None  \\
        BEVFusion \cite{liu2022bevfusion} & 71.4 / 68.5 & None  \\
        UVTR \cite{li2022uvtr} & 70.2 / 65.4 & None \\
        CenterFusion \cite{nabati2021centerfusion} & None & 45.3 / 33.2 \\
        mmFUSION \cite{ahmad2023mmfusion} & 69.8 / 65.4 & None \\
        FUTR3D \cite{chen2023futr3d} & 68.3 / 64.5 & 51.1 / 39.9 \\
        \textbf{Ours} & \textbf{71.8} / \textbf{69.0} & \textbf{51.7} / \textbf{40.3} \\
        \bottomrule
        \end{tabular}
    \vspace{-1em}
\end{table}

\subsection{Ablation Study}

To verify the impact of our local featured attention modules, we performed ablation experiments on nuScenes validation set in the conditions without our proposed local featured attention sub-modules on LiDAR-camera and radar-camera tasks. Training is conducted for 20 epochs with AdamW optimizer with a learning rate as $1.0 \times 10^{-4}$ for LiDAR-camera and radar-camera tasks. The results are shown in Table \ref{tab:ablation}.

\begin{table}[h] 
\small
\begin{center}
\caption{Ablation study on our FADet using nuScenes validation set.}
\begin{tabular}{c|cc}
\toprule
Experiments & NDS (\%) & mAP (\%) \\
\midrule
\hline
LiDAR + cam w/o featured attention & 67.9 & 63.3 \\
Radar + cam w/o featured attention & 49.4 & 37.2 \\
\hline
\end{tabular}
\label{tab:ablation}
\end{center}
\end{table}

\section{Conclusion}

In this work, we proposed a multi-sensor 3D object detection framework based on local featured attention, named FADet. We demonstrated our local attention modules on LiDAR, camera and radar input and verified their effectiveness via the experiments on nuScenes dataset. Our FADet is able to provide more accurate and comprehensive detection in long-tail and complex scenes from LiDAR, camera and radar sensors than baseline method FUTR3D. The results showed that our framework achieved $71.8\%$ NDS and $69.0\%$ mAP on LiDAR-camera detection tasks, $51.7\%$ NDS and $40.3\%$ mAP on radar-camera detection tasks. Since currently LiDARs, cameras and radars are optimal solutions for autonomous driving perception tasks, FADet is an important milestone to achieve universal yet robust performance of prediction for multi-sensor detection stacks.

\newpage
\bibliographystyle{IEEEtran}
\bibliography{references}

\end{document}